\documentclass{article}
\usepackage{cite}
\usepackage{amsmath,amssymb,amsfonts}
\usepackage{algorithmic}
\usepackage{graphicx}
\usepackage{textcomp}
\usepackage[table]{xcolor}
\usepackage{subcaption}
\usepackage{hyperref}
\usepackage[ruled,vlined]{algorithm2e}
\usepackage[symbol]{footmisc}
\usepackage{multirow}

\hypersetup{
    colorlinks=true,
    linkcolor=red,
    filecolor=magenta,      
    urlcolor=cyan,
}

\usepackage{geometry}
\usepackage{titlesec}

\makeatletter
\def\@seccntformat#1{\@ifundefined{#1@cntformat}%
  {\csname the#1\endcsname\quad}
 {\csname #1@cntformat\endcsname}}
\makeatother

\titleformat*{\section}{\Large\bfseries}
\titleformat*{\subsection}{\large\bfseries}
\titleformat*{\paragraph}{\large\bfseries}

\usepackage{times}

\usepackage[table]{xcolor}
\definecolor{lightgray}{rgb}{0.95, 0.95, 0.95}
\definecolor{darkgray}{rgb}{0.85, 0.85, 0.85}

\begin{document}
\twocolumn

\title{ \vspace{-12mm} \line(1,0){450} \\
\vspace{2mm}
Complex Human Action Recognition in Live Videos \\Using Hybrid FR-DL Method 
\\ \vspace{-2mm} \line(1,0){450}  \vspace{-6.7mm} \\\line(1,0){450}
}
\date{}

\vspace{-2mm}
\author{Fatemeh Serpush $^1$, \,\, Mahdi Rezaei $^{2,\, \star}$\\
$^1$ Faculty of Computer and Electrical Engineering, Qazvin Azad University, IR\\
$^2$ Faculty of Environment, Institute for Transport Studies, The University of Leeds, UK\\
$^1$ \href{mailto:f.serpush@qiau.ac.ir}{f.serpush@qiau.ac.ir}, \quad $^2$ \href{mailto:m.rezaei@leeds.ac.uk}{m.rezaei@leeds.ac.uk}}

\makeatother
\maketitle

\vspace{-1mm}
\textbf{Abstract:} Automated human action recognition is one of the most attractive and practical research fields in computer vision, in spite of its high computational costs. In such systems, the human action labelling is based on the appearance and patterns of the motions in the video sequences; however, the conventional methodologies and classic neural networks cannot use temporal information for action recognition prediction in the upcoming frames in a video sequence. On the other hand, the computational cost of the preprocessing stage is high. In this paper, we address challenges of the preprocessing phase, by automated selection of representative frames among the input sequences. Furthermore, we extract the key features of the representative frame rather than the entire features. We propose a hybrid technique using background subtraction and HOG, followed by application of a deep neural network and skeletal modelling method. The combination of a CNN and the LSTM recursive network is considered for feature selection and maintaining the previous information, and finally a Softmax-KNN classifier is used for labelling human activities. We name our model as \textit{``Feature Reduction \& Deep Learning''} based action recognition method, or FR-DL in short. To evaluate the proposed method, we use the UCF dataset for the benchmarking which is widely-used among researchers in action recognition research. The dataset includes 101 complicated activities in the wild. Experimental results show a significant improvement in terms of accuracy and speed in comparison with six state-of-the-art articles. \\

\vspace{-1mm}

\footnote[0]{$^\star$ Corresponding Author: \href{mailto:m.rezaei@leeds.ac.uk}{m.rezaei@leeds.ac.uk} (M. Rezaei)}

\vspace{-4mm}
\section{Introduction}
\vspace{-2mm}
Although the Human Activity or Action Recognition (HAR) is an active field in the present era, there are still key aspects which should be taken into consideration in order to accurately realise how people interact with each other or while using digital devices \cite{gammulle2017a}, \cite{hegde2017a}, \cite{ziaeefard2015a}. Human activity recognition is a sequence of multiple and complex sub-actions. This has been recently investigated by many researchers around the world using different type of sensors. Automatic recognition of human activities using computer vision has been more effective and a result with a growing demand in many applications. These include health care systems, activities monitoring in smart homes, Autonomous Vehicles and Driver Assistance Systems \cite{rezaei2014a}, \cite{rezaei2020a}, security and environmental monitoring to automatic detection of abnormal activities to inform relevant authorities about criminal or terrorist behaviours, services such as intelligent meeting rooms, home automation, personal digital assistants and entertainment environments for improving human interaction with computers, and even in the new challenges of social distancing monitoring during the COVID-19 pandemic \cite{azarmi2020a}. 

In general, we can obtain the required information from a given subject by using different types of sensors such as cameras and wearable sensors \cite{m2017a}, \cite{schneider2018a}, \cite{wu2016a}. Cameras are more suitable sensors for security applications (such as intrusion detection) and other interactive applications. By examining video regions, activities in different directions can be identified as forward or backward, rotation, or sitting positions. The concept of action and movement recognition in video sequences are very interesting and challenging research topics to many researchers. For example, in walking action recognition using computer vision and wearable devices, the challenges could be visual limitation of sensory devises. As a result, there may be a lack of available information to describe the movements of individuals or objects \cite{schneider2018a}, \cite{sharma2015a}, \cite{wang2016a}. 

On the other hand, the complex action recognition using computer vision demands a very high computation cost, while video capturing itself can be heavily influenced by light, visibility, scale, and orientation \cite{ullah2018a}. Therefore, in order to reduce the computational cost \cite{keyvanpour2019a}, a system should be able to efficiently recognise the subject's activities based on minimal data, given that the action recognition system is mostly online 
and needs to be assessed in real-time, as well. Accordingly, useful frames and frame index information can be exploited; In human pose estimation, the body pose is represented by a series of directional rectangles. Combination of rectangles' directions and positions defines a histogram to create a state descriptor for each frame. In the background subtraction methods (BGS), the background is considered as the offset and the methods such as histogram of oriented gradients (HOG), histogram of optical flow (HOF) and motion boundary histogram (MBH) can increase the efficiency of video based action recognition systems \cite{wang2018a}, \cite{patel2018a}. Skeletons models can capture the position of the body parts or the human hands/arms to  be used for human activity classification \cite{patel2018a}, \cite{khaire2018a}, \cite{shahroudy2018a}. Different machine learning methods have been proposed for action recognition and address the mentioned challenges, each of which has its own strengths, deficiencies and weaknesses. 

Convolutional Neural Networks (CNNs) is a type of deep neural network that effectively classifies the objects using a combination of layers and filtering \cite{varior2016a}. Recurrent Neural Network (RNN) can be used to address some of the challenges in activity recognition. In fact, RNNs include a recursive loop that retains the information obtained in the previous moments. The RNN only maintains a previous step that is considered as a disadvantage. Therefore, LSTM was introduced to maintain information of several sequential stages \cite{donahue2015a}. Theoretically, RNNs should be able to maintain long-term dependencies in order to solve two common problems of Exploding and Vanishing Gradients, while LSTM deals with the above-mentioned issues more efficiently \cite{m2017a}, \cite{ullah2018a}, \cite{ma2016a}, \cite{jain2016a}.

In the following sections we discuss in more details and provide further information about our ideas. The rest of this paper is organised as follows: Section~\ref{rel} reviews some of the most related work in the field. Section~\ref{met} explains the proposed method and procedures. In Section~\ref{exp} we will review the experimental and evaluation results and compare them with six state-of-the-art methods. Finally, Section~\ref{conc} concludes the paper and provides suggestions for future works.

\vspace{-1mm}
\section{Related Work}\label{rel}
\vspace{-2mm}
In the last decade, Human Action Recognition (HAR) has attracted the attention of many researchers from different disciplines and for various applications. Most of the existing methods use hand-crafted features, and thanks to the GPU and extended memory developments, the deep neural networks can also recognise the activities of subjects in the live videos. Human action recognition in a sequence of image frames is one of the research topics of the machine vision that focuses on correct recognition of human activities using single view images\cite{ullah2018a}, \cite{singh2019a}. In conventional hand-crafted approaches, the low-level features associated with an specific action were extracted from the video signal sequences, followed by labelling by a classifier, such as K-Nearest Neighbour (KNN), Support Vector Machine (SVM), decision tree, K-means, or Hidden Markov Models (HMMs) \cite{ullah2018a}, \cite{zolfaghari2018a}. Handcraft-based techniques require an expert to identify and define features, descriptors, and methods of making a dictionary to extract and display the features. Deep learning techniques for image classification, object detection, HAR, or sound recognition have also taken traditional hand-crafting techniques, but in a more automated manner than conventional approaches  \cite{sargano2017a}. \\

In \cite{ullah2018a}, the authors performed an analytical study on every six frames of input video sequences and tried to extract relevant features for action recognition using a pre-trained AlexNet Network. The method uses deep LSTM with two forward and backward layers to learn and extract the relevant features from a sequence of video frames. 

In \cite{sargano2017a}, a pretrained deep CNN is used to extract features, followed by the combination of SVM and KNN classifiers for action recognition. A pre-trained CNN on a large-scale annotation dataset can be transmitted for the action recognition with a small training dataset. So transfer learning using deep CNN would be a useful approach for training models where the dataset size is limited. By increasing the size of the dataset, the issue of overfitting will be eliminated; however, providing a large amount of annotated data is very difficult and expensive. In such conditions, the transfer learning is appropriate. The proposed technique in \cite{sargano2017a} aims to build a new architecture using a successful pre-trained model. 

In some research works, the human activity and hand gesture recognition problems are investigated using 3-D data sequence of the entire body and skeletons. Also, a learning-based approach, which combines CNN and LSTM, is used for pose detection problems and 3-D temporal detection \cite{ullah2018a}. Singh et al. \cite{singh2019a} propose a framework for background subtraction (BGS) along with a feature extraction function, and ultimately they use HMMs for action recognition. 

In \cite{patel2018a} an action recognition system is presented using various feature extraction fusion techniques for UCF dataset. The paper presents six different fusion models inspired by the early fusion, late fusion, and intermediate fusion schemes \cite{ghahroudi2007a}. In the first two models, the system utilises an early fusion technique. The third and fourth models exploit intermediate fusion techniques. In the fourth model, the system confront a kernel-based fusion scheme, which takes advantage of a kernel based SVM classifier. In the fifth and sixth models, late fusion techniques has been demonstrated. 

\cite{zolfaghari2018a} has processed only one frame of a temporal neighbourhood efficiently with a 2-D Convolutional architecture in order to capture appearance features of the input frames. However, to capture the contextual relationships between distant frames, a simple aggregation of scores is insufficient. Therefore, they feed the feature representations of distant frames into a 3-D network that learns the temporal context between the frames, so, it can improve significantly 
over the belief obtained from a single frame especially for complex long-term activities.

\cite{rahmani2018a} has proposed a Robust Non-Linear Knowledge Transfer Model (R-NKTM)
for human action recognition from unseen viewing angles. 
The proposed R-NKTM is a fully-connected deep neural network that transfers knowledge of human actions from any unknown view to a shared high-level virtual view by finding a non-linear virtual path that interconnects different views together. 
The R-NKTM is trained by dense trajectories of synthetic 3-D human models fitted to capture real motion data, and then generalise them for real videos of human actions. The strength of the proposed technique is that it trains only one single R-NKTM for all action detections and all viewpoints for knowledge transfer of any human action video, without the requirement of re-training or fine-tuning the model.
 
In \cite{lohit2018a} a probabilistic framework is proposed to infer the dynamic information associated with a human pose. The model develops a data driven approach, by estimating the density of the test samples. The statistical inference on the estimated density provides them with quantities of interests, such as the most probable future motion of the human and the amount of motion information conveyed by a pose. 

\cite{ullah2018a} proposes a novel robust and efficient human activity recognition scheme called ReHAR which can be used to handle single person activities and group activities prediction. First, they generate an optical flow image for each video frame. Then, both the original video frames and their corresponding optical flow images are fed into a single frame representation model to generate representations. Finally, an LSTM network is used to predict the forthcoming activities based on the generated representations.

\begin{figure}[t!]
\centering
\includegraphics[width=0.98\linewidth]{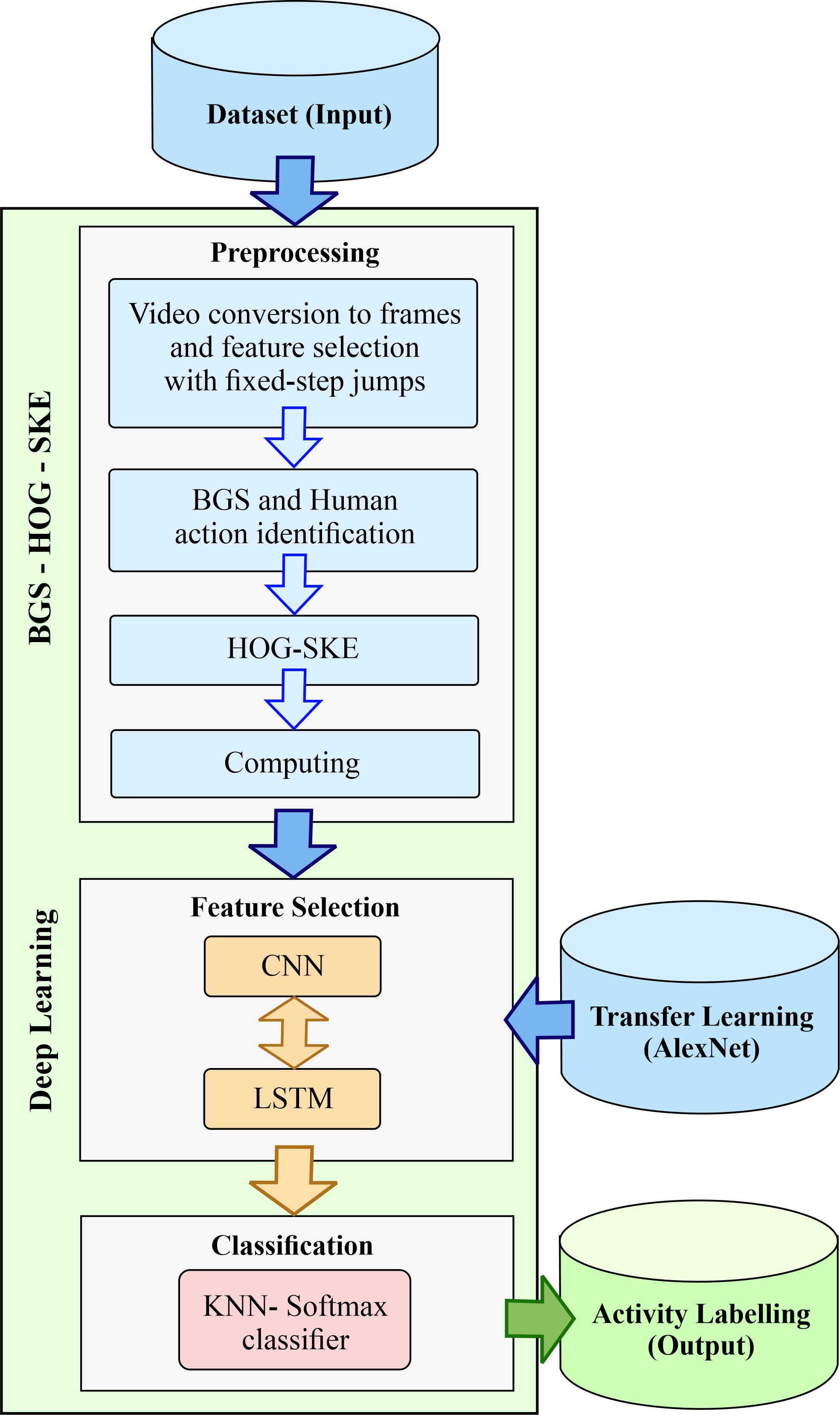}
\vspace{2mm}
\caption{The HAR system including two main modules: the preprocessing component and the deep neural network based feature extraction module FR--DL.}
\label{fig1}
\end{figure}
%

\vspace{-1mm}
\section{Methodology}\label{met}
\vspace{-2mm}
The architecture of our proposed method called Feature Reduction and Deep Learning (FR-DL) is shown in Figure \ref{fig1}. The proposed system consists of three main components: the input, the learning process, and the output. 

The learning process module includes Background Subtraction, Histogram of Oriented Gradients, and Skeletons (BGS-HOG-SKE), where we also call it feature reduction module; then we develope the CNN-LSTM model as deep learning module; and finally the KNN and Softmax layer as the human action classification sub-modules. The UCF101 dataset and AlexNet are also utilized in the system, the former is a collection of large and complex video clips and the latter one is a pre-trained system designed to enhance the system action detection performance. We use AlexNet for transfer learning \cite{ullah2018a} and as the backbone of the network. 

In the action recognition component, we have three sub-components: preprocessing, feature selection, and classifications. In the preprocessing step, the video clips are converted to a sequence of frames. However, the operations are performed only on selected frames which can have a positive impact on the cost and performance. Two deep CNN and LSTM neural networks are used to select the features with optimised weights. The parameters are trained on variety of datasets and are adjusted more precisely comparing to previous non-deep learning based methods. Later in section experimental we will show the main advantage of RNNs and deep LSTM with a higher accuracy rate in complex action recognitions, comparing to other deep neural network models. In the classification section, two methods of Softmax and KNN are used to label and classify the output as an action.\\

\begin{table}[t!]
\caption{The description of the Symbols}
\renewcommand\arraystretch{1.5}
\centering
\begin{tabular}{| l | l |}
\hline

\textbf{Symbols} & \textbf{Descriptions}\\ \hline \hline

\rowcolor{gray!20} J &  Jump length in input frames\\ 

$N_F$ & Number of representative frames\\ 

\rowcolor{gray!20}$f_k$ & $k^{th}$ representative frame\\ 

$v_i(u)$  & Value of pixel $u$ in block  $i$\\ 
     
\rowcolor{gray!20} & \\ 
\rowcolor{gray!20} \multirow{-2}{*}{$NG_i(u)$} & \multirow{-2}{13.5em}{\rule{0pt}{2.5ex}Spatial neighbourhood of pixel $u$ in block $i$} \\ 

 & \\
\multirow{-2}{*}{$sk=\{f_1,f_2,...,f_{N_F}\}$} & \multirow{-2}{13.5em}{\rule{0pt}{2.5ex}The sequence of skeleton with $N_F$ frames}\\ 

\rowcolor{gray!20}		& \\
\rowcolor{gray!20} \multirow{-2}{*}{$p^l = [p_1,...,p_I]$} & \multirow{-2}{13.5em}{\rule{0pt}{2.5ex}Output feature vector of the deep network \& input of classifications}\\ 

V & Video activity\\ 

\rowcolor{gray!20}TF(.) & Conversion function\\ \hline 

\end{tabular}
\label{tab1}
\end{table}

After the training phase of the developed action recognition system, the second phase is the system test and performance analysis, which specifies the system error and its accuracy. In Section \ref{exp} (Experimental results) we provide further details. 

Before we dive into the further technical details let's review on common symbols used in the following section. Table \ref{tab1} describes the symbols and notations used in this article.

\subsection{Learning Processing}
In learning process we have three stages of preprocessing, feature selection, and classification. The preprocessing stage is a very sensitive stage and the model performance highly depends on this stage and can lead to increased accuracy in the HAR output. In the following sub-sections more steps and details will be described.

\subsubsection{preprocessing}
\begin{figure}[t!]
	\centering
		\includegraphics[width=0.8\linewidth]{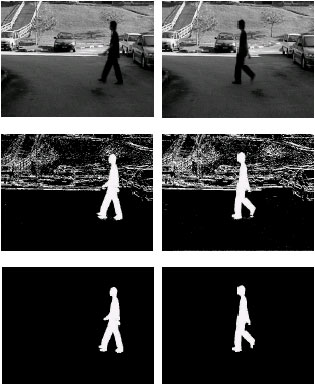}
	\caption{A sample representation of BGS technique to preprocess the extracted frames from a video.}
	\label{fig2}
\end{figure}

As shown in Figure \ref{fig2}, in the preprocessing stage, the input videos are converted into a sequence of frames. Then the representative frames will be selected from the given sequences of frames. In this study, we removed the background of representative frames using BGS technique (Figure \ref{fig2}, bottom row). After that we apply the deep and skeletal method 
on the representative frames, where depth motion maps explicitly create the motion representations from the raw frames. Below we explain our model in an step-by-step manner:\\

\noindent \textbf{A) Video to frame conversion and frame selection}: The input videos must first be converted to a set of frames \cite{anuradha2019a}, each of which is represented by a matrix as shown in Eq.~\eqref{eq1}:

\begin{equation} \label{eq1}
f_k =\left[\begin{array}{ccccc} f_{11}  & f_{12}  & . & . & f_{1m} \\ f_{21} & f_{22}  & . & . & f_{2m}  \\ . & . & f_{ij} & . & . \\ . & {.} & {.} & {.} & {.} \\ f_{n1} & f_{n2} & {.} & {.} & f_{nm} \end{array}\right]
\end{equation}

\noindent where $f_k$ is the $k^{th}$ representative frame, which has $n$ rows and $m$ columns. $f_{ij}$ are the feature values (intensity of each pixel) for the corresponding frame $k$.
After converting a video to frames we face a high volume of still images and frames that decrease the overall efficiency of the system due to high computational cost. In order to cope with the issue, we propose a simple yet effective solution to remove the redundant images. This can be done by fixed-step jumps $J$ to eliminate similar sequential frames \cite{wang2017a}. Based on our experimental, selecting one frame in every six frames will not significantly reduce the quality of the system, but speed it up significantly. We discuss this in more details later in section experimental result. Therefore, instead of extracting all features of all frames, only $N_F$ frames \cite{patel2018a}, \cite{khaire2018a}, \cite{shahroudy2018a}, \cite{liu2019a} were used. This makes our CNN network to perform more efficiently for the next steps. \\

 \begin{figure*}[t!]
	\centering
		\includegraphics[width=0.8\linewidth]{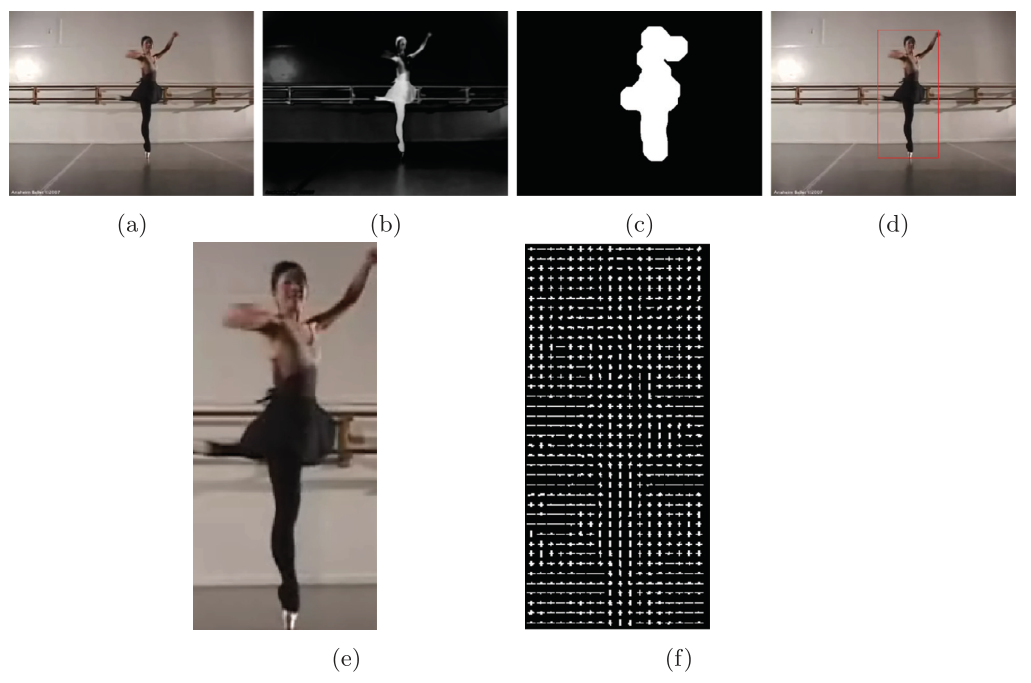}
	\caption{HOG steps for a sample \textit{``dancing''} action recognition \cite{patel2018a}}
	\label{fig3}
\end{figure*}

\noindent \textbf{B) BGS and human action identification}: A majority of the moving object recognition techniques include BGS, statistical methods, temporal differencing, and optical flow. The BGS scheme can be used indoors and outdoors, which is a popular method to separate moving parts of a scene by dividing it into background and foreground \cite{anuradha2019a}, \cite{dedeo2006a}, \cite{turaga2008a}. After separating the pixels from the static background of the scene, the regions can be classified into classes such as groups of humans. The classification algorithm depends on the comparison of the silhouettes of detected objects with pre-labelled templates in the database of an object silhouette. The template database is created by collecting samples of object silhouettes from samples of videos, labelled in appropriate categories. The silhouettes of the object regions are then extracted from the foreground pixel-map by using a contour tracing algorithm \cite{anuradha2019a}, \cite{dedeo2006a}. In \cite{wang2015a}, the BGS steps are described, where $f_k$ is the representative frame of the sequence of the video, assuming the neighbouring pixels share a similar temporal distribution. Given the pixel located in $u$, in the $i^{th}$ block of the image, the value and spatial neighbourhood are identified by $v_i(u)$ and $NG_i(u)$, respectively. Therefore, the value of the background sample of the pixel $u$, with $b_{i,j}(u)$ is determined to be equal to $v$, which is randomly chosen in $NG_i(u)$ (representative frame), as shown in Eq.\eqref{eq2}:
\begin{equation}\label{eq2} 
	b_{i,j}(u) = v(u|u \in NG_i(u))  \quad j = 1, 2, ..., l. 
\end{equation}
Then $A_i$ the background model of the pixel $u$ can be initialised by the background model of all pixels in the $i^{th}$ block:
\begin{equation}\label{eq3} 
A_{i} = [\{b_{i,1}(u)\} \{b_{i,2}(u)\} ,..., \{b_{i,l} (u)\} ], \quad u \in \mbox{Block}i. 
\end{equation} 
This strategy can extract the foreground of selected frames from short video sequences or from embedded devices with limited memory and processing resources. Additionally, minimal but efficient size of data is preferred as too large data sizes may result in statistical correlation destruction within the pixels in different locations. Further information for tracking the foreground extraction steps can be found in \cite{wang2015a}. This will also cause the difference between the intensity of each pixel in the current image decreases from the corresponding value in the reference background image.\\

An example of a sequence of BGS steps for walking is shown in Figure \ref{fig2}. Human shape plays an important role in recognising human action, which can extract blobs from BGS as shown in Figure \ref{fig2}, middle and bottom rows. Several methods based on global features, boundary, and skeletal descriptors have been proposed to illustrate the human shape in a scene\cite{turaga2008a}. After applying the BGS, a series of noise may disappear; however, some other noise may arise in other regions \cite{ke2013a}, \cite{tao2018a}. To remove such artefacts we use erosion and dilation morphological operators, with the structural elements of $3 \times 3$. The feature extraction step determines the diagnostic information needed to describe the human silhouette. In general, we can say that BGS extracts useful features from an object that increases the performance of our model by decreasing the size of the initial raw data, while maintaining the important parts of the embedded information.\\

\noindent \textbf{C) HOG-SKE: Histogram of Oriented Gradients and Skeleton}: In our proposed method, four different methods are used to evaluate the performance of the position descriptor: frame voting, global histogram, SVM classification, and dynamic time deviation. After that, the human body is extracted using complex screws or volumetric models such as cones, elliptical cylinders, and spheres. 
The HOG is a well-known feature extraction technique. HOG features can be extracted from the silhouette we made from the BGS stage, as also shown in Figure \ref{fig3} \cite{patel2018a}. The technique is a window-based descriptor used to compute points of interest, where the window is divided into an $n \times n$ frequency grid of the histograms. The frequency histogram is generated from each grid cell to indicate the magnitude and direction of the edge for every individual cell \cite{bajaj2019a}. 

 \begin{figure*}[t!]
	\centering
		\includegraphics[width=0.95\linewidth]{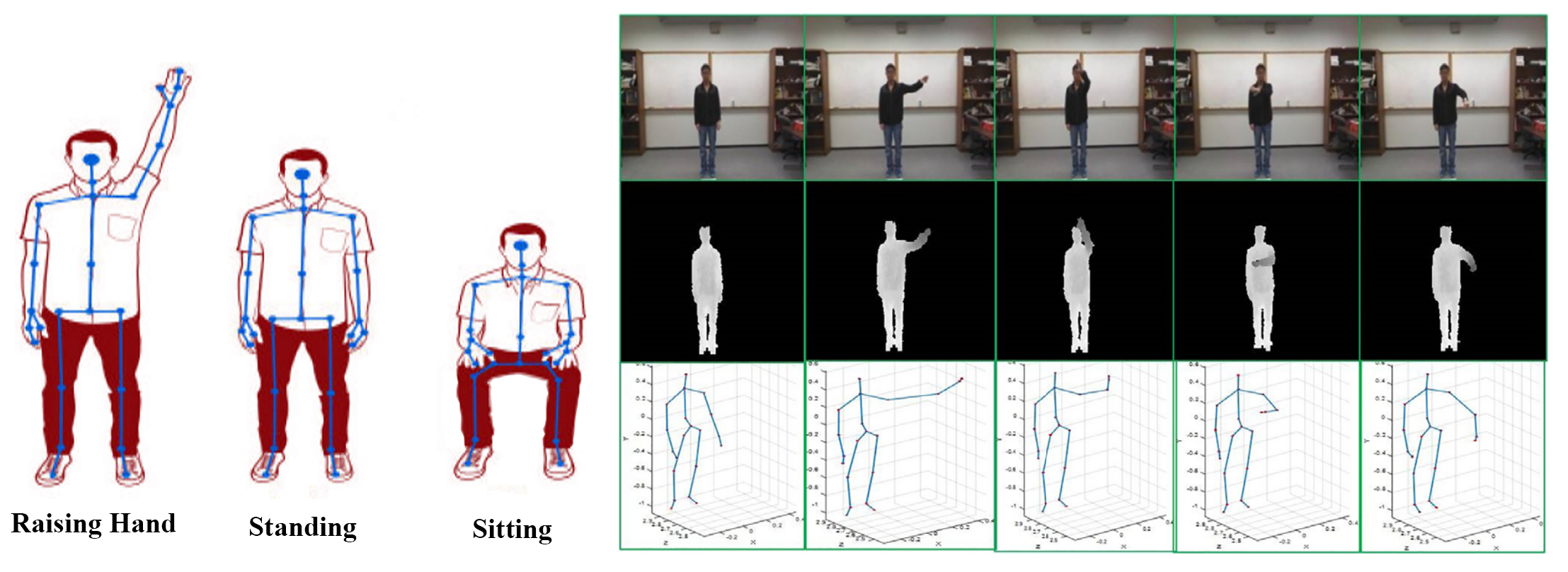}
	\caption{The steps of the appropriate frame  region selection and extraction of the skeleton motion \cite{khaire2018a}, \cite{pham2018a}}
	\label{fig4}
\end{figure*}

\noindent The cells are interconnected and HOG calculates the derivative of each cell (or sub-image), $I$, with respect to $X$ and $Y$ as shown in Eq.\eqref{eq4} and Eq.\eqref{eq5}:

\begin{equation}\label{eq4}
	I_X = I \times DX
\end{equation}

\noindent where $DX = [+1 \hspace{5mm}0 \hspace{4mm}-1]$,

\begin{equation}\label{eq5}
	I_Y = I \times DY
\end{equation}

\noindent where $DY = \left[\begin{array}{c} {+1} \\ {0} \\ {-1} \end{array}\right]$\\

$I_X$ and $I_Y$, are the derivative of the image with respect to $X$ and $Y$, respectively. In order to obtain these derivatives, horizontal and vertical Sobel filters (i.e. $DX$ and $DY$) are convolved on the image.

Normally, every video consists of hundreds of frames, and using the HOG will lead to an elongated vector and therefore a higher computational cost. For resolving these challenges, an overlap and 6 step frame jumps are used. 

Then magnitude and the angle of each cell is calculated as per the Eq.\eqref{eq6} and Eq.\eqref{eq7}, receptively. Finally  histograms of cells will be normalised.

\begin{equation}\label{eq6}       
|G| = \sqrt{I^2_X +\, I^2_Y}	
\end{equation}
\begin{equation}\label{eq7}       
\phi= \arctan \left(\frac{I_X}{I_Y}\right)
\end{equation}

\vspace{2mm}
\noindent In this paper, in addition to the HOG method a simple skeleton view is also used for action recognition. Real-time Skeleton estimation algorithms are used in commercial deep integrate cameras. This technology allows the fast and easy joints extraction of human body \cite{bajaj2019a}. Some studies, only use part of the body in a skeleton method, such as hands. However, in this research, the whole body is used to increase the overall accuracy. Figure \ref{fig4}-left illustrates a skeletal method on three activities of sitting, standing, and raising hand and Figure \ref{fig4}-right focuses more on hand activity recognition. 

One of the advantages of deep data and skeletal data, as compared with traditional RGB data is that, they are less sensitive to changes in lighting conditions \cite{khaire2018a}. We use Skeleton and inertia data at both levels of feature and decision making to improve the accuracy of our action recognition model.

The sequences $s_k$ of the skeleton with $N_F$ frames are shown as: $s_k = \{ f_1 ,f_2 ,...f_{N_F} \}$. We use same notations as in \cite{pham2018a}. 

To represent spatial and temporal information, the coordinate skeleton sequence $(X_i, Y_i, Z_i)$ is considered. For each $f_i$ skeleton, $i\in [1, N_F]$, in the range [0, 255], and the normalisation operation is performed according to the Eq.\eqref{eq8} with the TF(.) conversion function:
\begin{equation}\label{eq8} 
\begin{array}{l} 
{(X_i^{'} ,Y_i^{'} ,Z_i^{'}) = TF(X_i, Y_i, Z_i)} \\ 
{} \\ 
X_i^{'} = 255 \times \frac{X_i - \min \{C\}} {\max \{C\} - \min \{C\}} \\
 \\ 
Y_i^{'} = 255 \times \frac{Y_i - \min \{C\}} {\max \{C\} - \min \{C\}} \\ 
\\ 
Z_i^{'} = 255 \times \frac{Z_i - \min \{C\}} {\max \{C\} - \min \{C\}} \\
\end{array} 
\end{equation} 

\vspace{2mm}
\noindent where $\min\{C\}$ and $\max\{C\}$ are minima and maxima of all coordinate values.\\

\begin{figure*}[t!]
	\centering
		\includegraphics[width=0.97\linewidth]{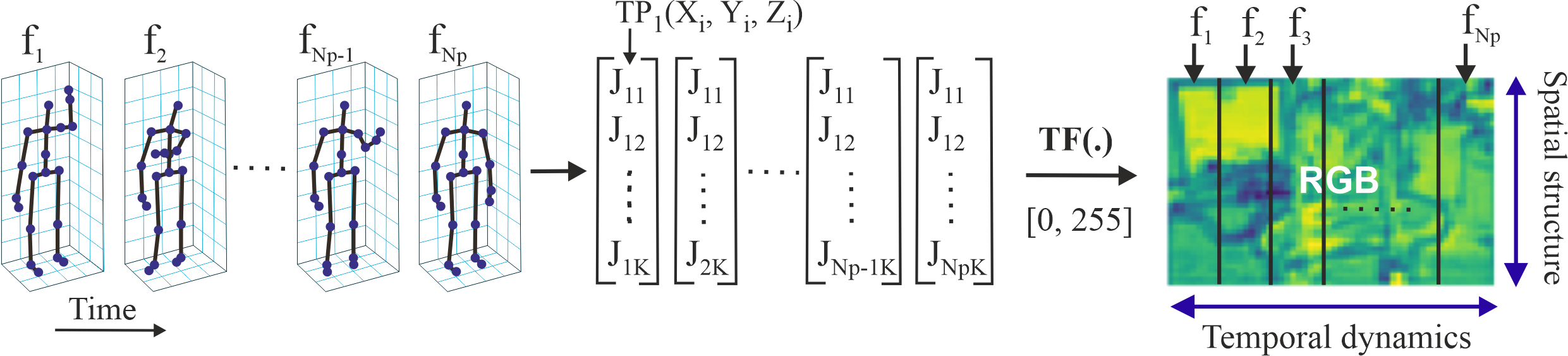}
	\caption{The stages of converting skeletal sequences to spatio-temporal information to train the model.}
	\label{fig5}
\end{figure*}

The new coordinate space is quantified to integral image representation and three coordinates $(X'_i, Y'_i, Z'_i)$ are considered as the three components R, G, B of a colour-pixel:\\
\begin{equation*}
	(X'_i=R,\, Y'_i=G,\, Z'_i=B)
\end{equation*}
 
$(X'_i, Y'_i, Z'_i)$ is the new coordinate of the image display. The steps are shown in Figure \ref{fig5}. Following the above steps and conversions, the raw data of the skeleton sequence changes into 3-D tensors and then is injected into the learning model as inputs. In Figure \ref{fig5}, $F_N$ denotes the number of frames in each skeleton sequence. $K$ denotes the number of joints in each frame and it depends on the deep sensors and data acquisition settings.\\

\noindent \textbf{D) ROI Calculation}: During the process of feature extraction to display action, a combination of contour-based distance signal feature, flow-based motion feature \cite{ullah2018a}, \cite{wang2018a}, and uniform rotation local binary patterns can be used to define region of interest for feature extraction \cite{patel2018a}, \cite{khaire2018a}, \cite{shahroudy2018a}, \cite{singh2019a}. Therefore, at this stage, suitable regions for extraction of the feature are determined. Depending on the nature of the dataset, the input videos may include certain multi-view activities, which increase the accuracy of the classification. A similar method is presented in \cite{rahmani2018a}, \cite{kumari2017a} for extraction of Entropy-based silhouettes.

\subsubsection{Feature selection}
Given that in each movie an action is represented by a sequence of frames, we can perform the action recognition by analysing the contents of multiple frames in a sequence. We propose a series of techniques and methods to find out activities that are close to human perceptions of activities in real life. 

One of the human abilities is to predict the upcoming actions based on the previous action sequences. Therefore, to enable a system with such characteristics, deep neural networks, inspired from natural human neural networks is very appropriate. These networks include but not limited to CNN, RNN, and LSTM. 

In many research works, the CNN streams are fused with RGB frames and skeletal sequences at feature level and decision level. Classification at decision-making level is also done through voting strategy. As already mentioned, the existence of multidimensional visual data encourages us to combine all vision cues, such as depth and skeletal as in\cite{khaire2018a}. Many studies focus on the improved skeletal display of CNN architecture. One of the major challenges in exploiting CNN-based methods for detecting skeletal-based action is how to display a temporal skeleton sequence effectively and feed them into a CNN for feature learning and classifications. 

To overcome this challenge, we encode the temporal and spatial dynamics of skeleton sequences in 2-D image structures 
CNN is used to learn the features of the image and its classification to identify the original skeleton sequences \cite{n2018a}. CNN generally consists of convolutional layers, pooling layers and fully-connected layers. In the convolutional layer, filters are very useful for detecting the edges in the images \cite{ronao2015a}, \cite{chavarriaga2013a}, \cite{wu2016b}, \cite{zeng2014a}, \cite{wang2017b}. The pooling layers are generally used in the Max-type, which is intended to reduce the dimension, and the fully-connected layers are used to convert a cubic dimensional data in to a 1-D vector \cite{montes2016a}. 

Based on a stack of $N_F$ input frames, this convolutional network learns to optimise the filters weight; however, it may not be capable of detecting complicated video sequences with complex activities, such as eating or jumping over obstacles. RNNs can resolve this problem \cite{ullah2018a}, \cite{molchanov2016a}, \cite{mahasseni2016a}, by storing only the previous step and consequently avoiding the exploding and vanishing gradient issue. It can be said that the LSTM network is a kind of RNN, which solves the aforementioned issues by holding up a short memory for a long time. In our research, we combine CNN and LSTM for feature selection and accurate action recognition due to their high performance in visual and sequential data. AlexNet is also injected into feature selection for identifying hidden patterns of the visual data. The feature selection operation is performed in parallel in order to speed up the processing, namely, parallel duplex LSTMs. A similar approach is considered in \cite{ullah2018a}, \cite{zolfaghari2018a}, \cite{li2018a}, \cite{huan2019a}, and \cite{park2016a}. In other words, we use LSTM for two main reasons:

\begin{enumerate}
\item  As each frame plays an important role in a video, maintaining the important information of successive frames for a long time will make the system more efficient. The "LSTM" method is appropriate for this purpose.

\item  Artificial neural networks and LSTM have greatly gained success in the processing of sequential multimedia data and have obtained advanced results in speech recognition, digital signal processing, image processing, and text data analysis \cite{ullah2018a}, \cite{n2018a}, \cite{zhu2017a}. 
\end{enumerate}

\begin{figure*}[t!]
	\centering
		\includegraphics[width=0.7\linewidth]{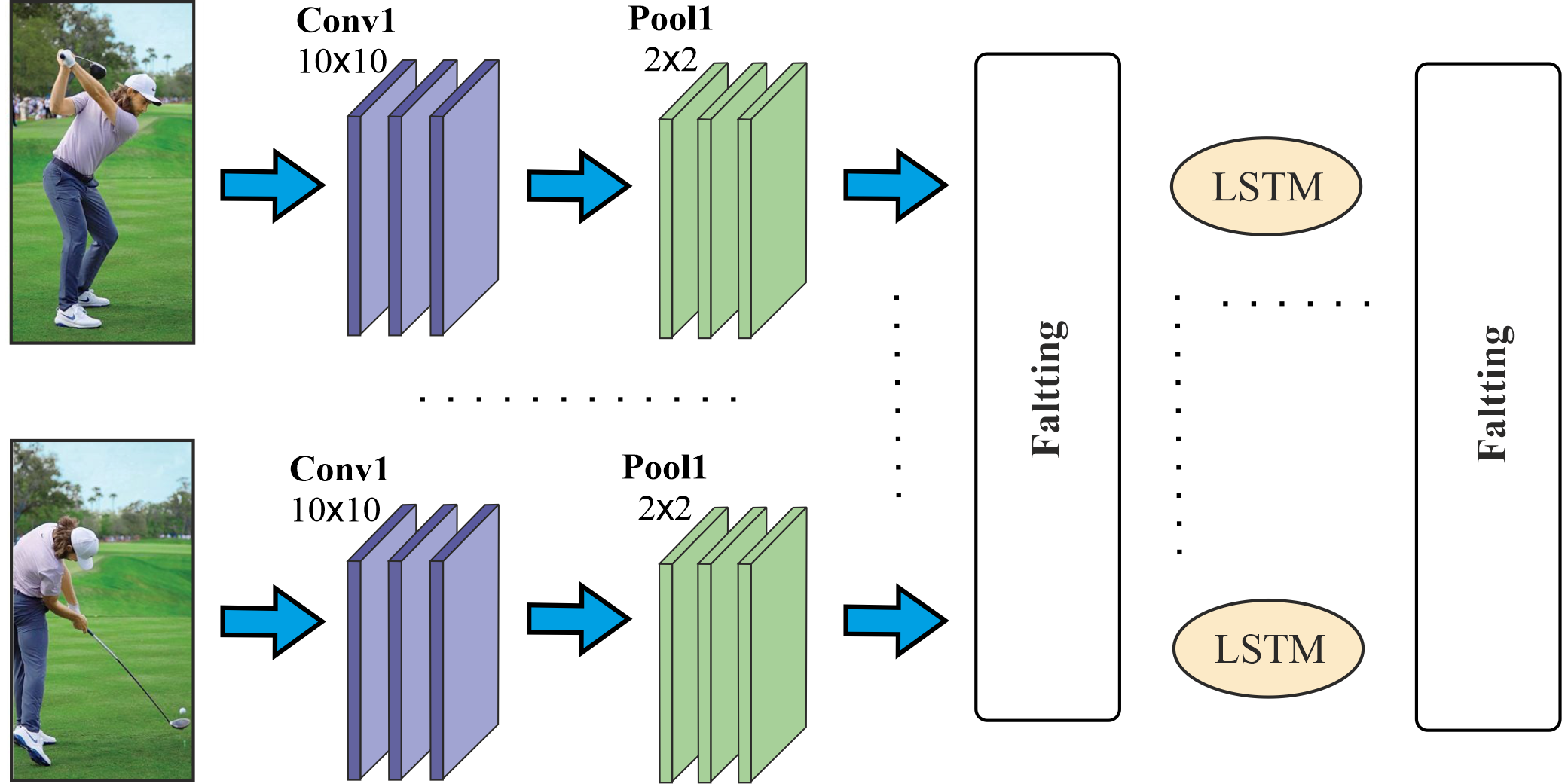}
	\caption{The simplified architecture of the proposed Deep Learning model based on hybrid CNN and parallel LSTM to select the deep features of a given frame set (e.g. Golf swing action recognition using spatio-temporal information)}
	\label{fig6}
\end{figure*}

Figure~\ref{fig6} describes how we use a CNN and dual LSTM networks in our work. According to research conducted in \cite{ullah2018a}, \cite{montes2016a}, \cite{wang2018b}, \cite{pham2018a}, \cite{ullah2018b}, LSTM is capable of learning long-term dependencies, and its special structure includes inputs, outputs and forget gates, which controls long-term sequence recognition. The gates are set by the Sigmoid unit opened and closed during the training. Each LSTM unit is calculated as Eq. \eqref{eq9} to \eqref{eq15}:

\vspace{-1.5mm}
\begin{equation}\label{eq9}    
i_{t} = \sigma ((x_{t} +s_{t-1} ) W^i + b_i)
\end{equation}

\vspace{-1.5mm}
\begin{equation}\label{eq10}    
f_{t} = \sigma ((x_{t} +s_{t-1} ) W^f + b_f)
\end{equation}

\vspace{-1.5mm}
\begin{equation}\label{eq11} 
o_{t} = \sigma ((x_t +s_{t-1})W^o + b_o)
\end{equation}

\vspace{-1.5mm}
\begin{equation}\label{eq12} 
g = \tanh((x_t + s_{t-1}) W^g + b_g)
\end{equation}

\vspace{-1.5mm}
\begin{equation}\label{eq13}
c_t = c_{t-1} \odot f_t + g \odot i_t
\end{equation}

\vspace{-1.5mm}
\begin{equation}\label{eq14}  
s_t = \tanh (c_t)\odot o_t
\end{equation}

\vspace{-1.5mm}
\begin{equation}\label{eq15}
Final state = Softmax(Vs_t)
\end{equation}

\noindent where $x_t$ is the input at time $t$, $f_t$ is the forget gate at time $t$ which clears the information from the memory cell, if needed, and holds a record of the previous frame. Output gate $o_t$ holds the information about the next step, $g$ is the return unit and has the \textit{tanh} activation function which is computed using the current frame input and the previous $s_{t-1}$ frame status. $s_t$ is the RNN output from the current mode. The hidden mode is calculated from one RNN stage by activating \textit{tanh} and $c_t$ memory cells. $W^i$ is the input gate weight, $W^o$is the output gate weight, $W^f$ is the forget gate weight and $W^g$ is the returning unit weight from the LSTM cell. $b_i$, $b_o$, $b_f$ and $b_g$ are the biases for input, output, forget and the returning unit gates, respectively.

 As the action recognition does not need the intermediate output of the LSTM, we made a final decision making by applying a Softmax classifier on the final state of the RNN network. Training large data with complex sequence patterns (such as video data) can not be identified by a single LSTM cell, so we use stacking multiple LSTM cells to learn long term dependencies in video data.

%
%
\subsection{Transfer learning: AlexNet}
AlexNet is an architecture for solving the challenges of the human action recognition system, trained on the large ImageNet dataset with more than 15 million images. The model is able to identify hidden patterns in visual data more accurately than many other CNN based architectures \cite{ullah2018a}, \cite{sargano2017a}. Action recognition system requires high training data and computing ability. AlexNet is embedded in the architecture of our model to extract the higher-performing features because the pre-trained AlexNet does not have any negative impacts on the performance of the system. 

The AlexNet architectural parameters are presented in Table \ref{tab2}. It has six layers of convolution, three layers of pooling and three fully-connected layers. Each layer is followed by a non-linear ReLU activation function and the vector of extracted features from the FC8 layer is 1000-dimensional.\\

\subsubsection{KNN-Softmax classifier}
Classification is usually done in deep neural networks based on Softmax function. The Softmax classifier practically is placed after the last layer in the deep neural network. In fact, the result of the convolutional and pooling layers (a feature vector $p^l = [p_1, ..., p_I]$) is the input of the Softmax \cite{chavarriaga2013a}, \cite{zeng2014a}. After forward propagation, weighs are updated, and errors are minimised through an stochastic gradient descent (SGD) optimisation on several training examples and iterations. Back-propagation balances the weights by calculating the the gradient of convolution weights. In case of large number of classes the Softmax does not perform very well. This is normally due to two main reasons: when the number of parameters is large, the last layer fails to increase the forward-backward speed; furthermore, syncing GPUs will be difficult as well \cite{zeng2014a}, \cite{wang2017b}.\\

\begin{figure*}[t!]
	\centering
		\includegraphics[width=0.8\linewidth]{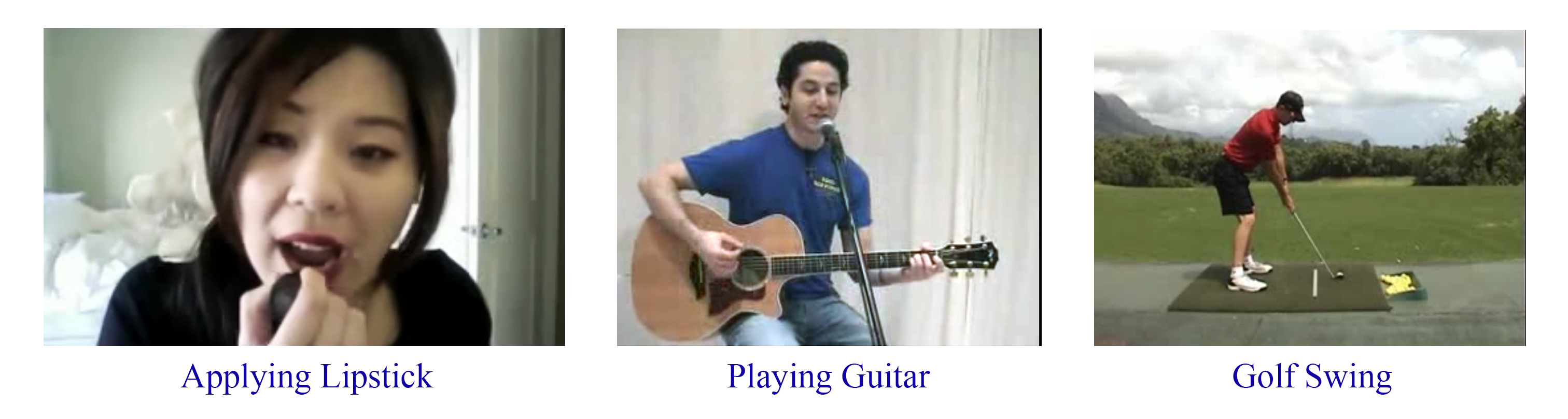}
		\vspace{-4mm}
	\caption{Sample frames and activities from the UCF 101 video dataset}
	\label{fig7}
\end{figure*}

\begin{table*}[t]
\caption{AlexNet Architecture Specifications}
\vspace{-2mm}
\renewcommand\arraystretch{1.5}
\centering
	\begin{tabular}{ l | c | c | c | c | c | c | c | c | c | c | c }
		\hline
		\textbf{Layers} & \textbf{Conv1} & \textbf{Pool1} & \textbf{Conv2} & \textbf{Pool2} & \textbf{Conv3} & \textbf{Conv4} & \textbf{Conv5} & \textbf{Pool5} & \textbf{FC6} & \textbf{FC7} & \textbf{FC8}\\ \hline \hline
		\textbf{Kernel} & $11 \times 11$ & $3 \times 3$ & $5 \times 5$ & $3 \times 3$ & $3 \times 3$ & $3 \times 3$ & $3 \times 3$ & $3 \times 3$ & - & - & - \\ \hline
		\textbf{Stride} & 4 & 2 & 1 & 2 & 1 & 1 & 1 & 2 & - & - & - \\ \hline
		\textbf{Channels} & 96 & 96 & 256 & 256 & 384 & 384 & 256 & 256 & 4096 & 4096 & 1000 \\ \hline
\end{tabular}
\label{tab2}
\end{table*}

In this article we use KNN when the number of classes is high and Softmax fails to perform well. After classifying by Sofmax, if it fails (that is the probability of closeness of action to two classes or several classes), then KNN should be used. KNN uses Euclidean distance \cite{sharif2019a} and Hamming distance to detect the similarity between two feature vectors. As previously mentioned, $p^l = [p_1,..., p_I]$ is a classifier input which holds\\ \vspace{-2mm}

\qquad \qquad $p_I = \{(x_i, y_j), \,\, i = 1, 2, ..., n_I \}$.\\ \vspace{-2mm}

\noindent where $x_i$ is the number of extracted features and $y_j$ is the equivalent label for each feature set of $x_i$. We use Euclidean distance in the KNN classifier, with $k = 10$ and squared inverse distance weights \cite{sharif2019a}. The Euclidean distance is formulated as Eq. \eqref{eq16}.

\begin{equation}\label{eq16} 
d(x_i ,\, x_i+1) = {\mathop{\min }\limits_i} \, (d(x_i ,\, x_{i+1})) 
\end{equation} 

Assuming $u$ is a new instance with a label $y_j$, in order to find $v+1$ and closest neighbour to $u$, the distance formula with $d(u,\, x_i)$ can be determined as Eq. \eqref{eq17}.

\begin{equation}\label{eq17}
d(u, x_i) = \frac {d(u,\, x_i)}{d(u,\, x_{v+1})}
\end{equation}

\noindent we normalise $d(u,x_{i} )$ by the kernel function and weighing according to Eq. \eqref{eq18}.
                            
\begin{equation}\label{eq18}
w(i) = k (d(u,\, x_i))
\end{equation}

\noindent The final membership function of weighted K-nearest neighbour (W-KNN) is formulated as follows:

\begin{equation}\label{eq19}
	\hat{y} = max_{j} \left( \sum_{i=1} w(i)I(y_i = j)\right)
\end{equation}

\section{Experimental Results}\label{exp}
In this section, we will evaluate the proposed method on the UCF101 dataset as a common benchmarking dataset based on the accuracy criterion, followed by discussion on the experimental results. The dataset is divided into three parts: training, testing, and validation, based on 60\%, 20\% and 20\% splits, respectively. In Figure \ref{fig7}, examples of dataset are shown. To implement the proposed model we use Python 3 and TensorFlow deep learning framework. 

\noindent In our evaluations, we compare the proposed method with six state-of-the-art methods using the accuracy criterion.

\begin{figure*}[t!]
	\centering
		\includegraphics[width=0.65\linewidth]{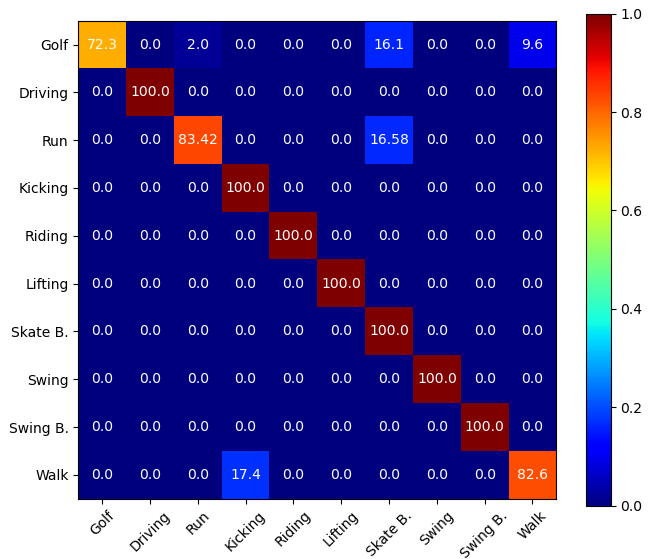}
		\vspace{-1mm}
	\caption{Confusion matrix of the proposed FR-DL method for sport actions}
	\label{fig8}
\end{figure*}
%

\subsection{UCF101 dataset}
The UCF101 dataset is a relatively complex dataset due to many action categories. Some categories include variety of actions, such as sport related actions. The videos are captured in different lighting conditions, gestures, and viewpoint. One of the major challenges in this dataset is the mixture of natural realistic actions and the actions played by many actors while in other datasets the activities and actions are usually performed by one actor only\cite{ullah2018a}, \cite{sargano2017a}, \cite{park2016a}, \cite{sharif2019a}. The UCF101 dataset includes 101 action classes, over 13,000 video clips and 27 hours of video data. This dataset contains realistic uploaded videos with camera motion and custom backgrounds. Therefore, UCF101 is considered as a very comprehensive dataset. The action categories in this dataset can generally be considered as five major types: Interaction between human and object, body movement, human-to-human interaction, musical instruments, and sport \cite{ullah2018a}, \cite{tu2018a}, \cite{soomro2012a}. Figure \ref{fig7} shows one sample frame of three different video clips and actions from the UCF101 dataset. Sport category is the largest category of the UCF101 dataset and plays an important role in benchmarking.

\subsection{UCF Sports dataset}
This dataset contains 150 videos of sports broadcasts that are captured in cluttered, and dynamic environments. There are 10 action classes and each video corresponds to one action \cite{tu2018a}. In some research works such as \cite{chaquet2013a} which are based on temporal template matching, the UCF Sports action has been used for benchmarking purposes. This category is also useful for actions that are related to human body motion such as ``Walk'' or to human-object interaction such as ``Horse-Riding'' \cite{tu2018a}.

In the next two sections we discuss about two types of the test and evaluations that we conducted in this research:

\begin{table}[t]
\caption{Performance evaluation on the proposed method comparing to seven other methods}
\renewcommand\arraystretch{1.5}
	\centering
	\begin{tabular}{l | c | c}
		\hline
		\textbf{Methods} & \textbf{Year} & \textbf{Accuracy \%}\\ \hline \hline
		DB-LSTM \cite{ullah2018a}        & 2018 & 91.22 \\ \hline
		SVM-KNN \cite{sargano2017a}      & 2017 & 91.47 \\ \hline
		R-NKTM \cite{rahmani2018a}       & 2018 & 90.00 \\ \hline 
    Lohit et al. \cite{lohit2018a}   & 2018 & 57.90 \\ \hline 
    ECO \cite{zolfaghari2018a}       & 2019 & 93.10 \\ \hline 
    Patel et al. \cite{patel2018a}   & 2018 & 89.43 \\ \hline 
    \textbf{FR-DL} (Proposed method) & 2020  & \textbf{93.90}\\ \hline
 \end{tabular}
\label{tab3}
\end{table}

\subsection{FR-DL Performance Evaluation}
Table \ref{tab3} presents the outcome of our experiments for the proposed FR-DL method. The proposed method shows an improvement rate of 0.8\% to 4.47\% comparing to six other state-of-the art method.

The proposed hybrid FR-DL method uses a combination of BGS, HOG and Skeleton methods as the preprocessing stage, to extract features that played a major role in the action recognition. The combination of convolution, pooling, fully-connected and LSTM units are used to achieve a better feature learning, feature selection, and classification. Therefore, the probability of error in the classification stage is greatly reduced; and furthermore, complex activities are recognised with higher accuracy rate, as well.  

\subsection{Optimum Frame Jumping}
\vspace{-1mm}
As the second experiment we also evaluated the optimum jump length for the proposed FR-DL method. Every video is considered as a single input, and then the features of the frames are extracted using one frame out of every $x$ frames. Table \ref{tab4} shows the evaluation of the proposed method, based on different frame jumps of 4, 6, and 8 and their impact on the performance of the system. Using the frame jump of $J=6$ we achieved nearly 50\% improvement in speed and computational cost of the system in comparison with $J=4$, while we approximately lost only 1.5\% in accuracy rate. Therefore, considering the speed-accuracy trade-off, we selected the frame jump of 6 as the optimum value for our intended application while it still outperforms the similar state-of-the-art method (DB-LSTM) \cite{ullah2018a}.

\begin{table*}[t]
\vspace{-1mm}
\caption{Evaluation of the proposed method based on different jumps}
\vspace{-2mm}
\renewcommand\arraystretch{1.3}
	\centering
	\begin{tabular}{ l | c | c | c}
		\hline
		\textbf{Methods} & \textbf{Frame Jump} & \textbf{Average time (S)} & \textbf{Average Acc.} \%\\ \hline \hline
		& 4.0 & 1.72 & 92.2\% \\ \cline{2-4}
		\multirow{3}{0em}\textbf{DB-LSTM} \cite{ullah2018a}    & 6.0 & 1.12 & 91.5\% \\ \cline{2-4}
		   & 8.0 & 0.9 & 85.34\% \\ \hline 
     & 4.0 & 2.10 & 95.62\% \\ \cline{2-4}
     \multirow{3}{0em}\textbf{FR-DL} (Proposed method) & 6.0 & 1.6 & 93.9\% \\ \cline{2-4}
       & 8.0 & 1.10 & 89.6\% \\ \hline 
 \end{tabular}
\label{tab4}
\end{table*}

\subsection{Confusion Matrix}
\vspace{-1mm}
A confusion matrix contains visualised and quantised information about multiple classifiers using a reference classification system \cite{ullah2018a}, \cite{huan2019a}. Each row represents the predicted class, and each column represents instances of the ground truth classes. Figure~\ref{fig8} shows the details of the results on the UCF Sports dataset for the proposed FR-DL method. 

The confusion matrix results confirms that in overall the FR-DL provides a more consistent confusion matrix comparing the ReHAR method \cite{ullah2018a}, even for Golf, Run, and Walk actions as our weakest results with the accuracy of 82.6\%, 83.42\%, and 72.30\%, respectively, in contrast to 83.33\%, 75.00\%, and 57.14\% for the ReHAR method.

As per Figure~\ref{fig8}, it can also be interpreted that we have examples of ``walking'', ``running'' and ``Golf''activities which are mistakenly identified as ``Kicking'',``Skateboarding'', and ``walking'', respectively. This are expectable, as some of these actions have common features that lead to a misclassification. Furthermore, extra objects and people in the background of the scene are among the factors that also leads to a wrong classification. For example, in one of the examined videos ``walk-front/006RF1-13902-70016.avi'', there is a person who walks on a golf field with a golf pole. The environment is related to golf field and the motion of the golf pole in the background looks like a person is swinging the pole in front of him \cite{huan2019a}. This was an examples of mis-classifications by the proposed FR-DL method.\\

\vspace{-3mm}
\subsection{Discussion}
According to summarised performance in Table \ref{tab3}, the proposed FR-DL method helped us in recognising complex actions using spatio-temporal information, which were hidden in sequential patterns and features. We initialised the weights randomly and trained all the networks by reiterating the training stage until getting the minimum errors \cite{pham2018a}, \cite{fernando2016a}, \cite{ma2019a}, \cite{singh2016a}, \cite{ye2018a}. In addition to Softmax, the KNN is used for classification. These together made the proposed method more precise than the other methods, as seen in Table \ref{tab3}.  

The system can reduce the effects of the degradation phenomenon for both training and test phases. It should be noted that degradation phenomena considerably depends on the size of the datasets. This is the reason why the networks with too many layers have higher errors than medium-size networks. The difference between the training error and the test error on the learning curves shows the ability of overfitting prevention. 

We also extended the skeleton encoding method by exploiting the Euclidean distance and the orientation relationship between the joints. According to Table \ref{tab4}, in both methods the accuracy level is slightly higher with $J=4$; however, the time complexity of Jump 6 is significantly less than the Jump 4. Therefore, the Jump 6 is considered as a better trade-off in terms of accuracy and time complexity. 

As the table \ref{tab4} shows, the DB-LSTM method is slightly faster than the FR-DL method but less accurate. In general, the suggested method of jumping 6 requires 1.6 second in a medium range Cori7 PC to process a 1-second og HD video clip \cite{ullah2018a}. Depending on the nature of the application in terms of speed and accuracy requirements, this can be simply converted to a 1-1 real-time action recognition solution either by increasing the jump step, or by improving the CPU speed, or reducing the input video resolution, or by considering a combination of all three factors. 

\vspace{-2mm}
\section{Conclusion}\label{conc}
\vspace{-2mm}
In this article, we proposed a new approach including a combination of BGS, HOG and Skeletal to analyse and describe the appropriate frames for the preprocessing phase of the human action recognition. Then a deep convolution neural network and LSTM for the feature selection were implemented, and finally the Softmax-KNN for the labelling and classification of the actions was utilised. The experimental was performed on the commonly used UCF dataset. The accuracy metric, time complexity, and confusion matrix were assessed and analysed, and the overall results showed the proposed FR-DL method outperforms in human action recognition compared with six other state-of-the-art research in the field. Action recognition has many applications, including in smart homes, surveillance systems, autonomous driving, etc. and has already attracted the attention of many researchers that can even remotely monitor people at work, in living environments, or in public places. As future work, we suggest extending this research to improve the current architecture and tp predict the future actions based on spatio-temporal information, current action, and semantic scene segmentation and understanding.
\vspace{-1mm}
\bibliographystyle{ieeetr}
\bibliography{biblio}

\end{document}